\documentclass[runningheads]{llncs}
\usepackage[T1]{fontenc}
\usepackage{graphicx}
\usepackage{amsmath} 
\usepackage{booktabs}

\usepackage{color,soul}
\usepackage[hidelinks]{hyperref}

\usepackage{wrapfig}

\begin{document}

\title{Knowledge Graph Structure as Prompt: Improving Small Language Models Capabilities for Knowledge-based Causal Discovery}

\titlerunning{\texttt{KG Structure as Prompt}: SLMs for Knowledge-based Causal Discovery}

\author{Yuni Susanti\inst{1}\textsuperscript{\orcid{0009-0001-1314-0286}}\thanks{This work was conducted during a research stay at KIT and ScaDS.AI/TU Dresden, Germany.} 
\and
Michael Färber\inst{2}\textsuperscript{\orcid{0000-0001-5458-8645}}}

\newcommand{\orcid}[1]{\href{https://orcid.org/#1}{\includegraphics[width=10pt]{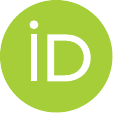}}}

\authorrunning{Y. Susanti and M. Färber}

\institute{Artificial Intelligence Lab., Fujitsu Ltd., Japan\\
\email{yuni.susanti@fujitsu.com}
\and
ScaDS.AI \& TU Dresden, Germany\\
\email{michael.faerber@tu-dresden.de}
}

\maketitle  
\setcounter{footnote}{0}

\begin{abstract}
Causal discovery 
aims to estimate causal structures among variables based on observational data. Large Language Models (LLMs) offer a fresh perspective to tackle the causal discovery problem by reasoning on the metadata associated with variables rather than their actual data values, an approach referred to as \textit{knowledge-based causal discovery}. In this paper, we investigate the capabilities of Small Language Models (SLMs, defined as LLMs with fewer than 1 billion parameters) with prompt-based learning for knowledge-based causal discovery. Specifically, we present ``\texttt{KG Structure as Prompt}'', a novel approach for integrating structural information from a knowledge graph, such as \textit{common neighbor nodes} and \textit{metapaths}, into prompt-based learning to enhance the capabilities of SLMs. 
Experimental results on three types of biomedical and open-domain datasets under few-shot settings demonstrate the effectiveness of our approach, surpassing most baselines and even conventional fine-tuning approaches trained on full datasets. Our findings further highlight the strong capabilities of SLMs: in combination with knowledge graphs and prompt-based learning, SLMs demonstrate the potential to surpass LLMs with larger number of parameters.
Our code and datasets are available on GitHub.\footnote{\url{https://github.com/littleflow3r/kg-structure-as-prompt}}

\end{abstract}

\section{Introduction}

\begin{wrapfigure}[13]{r}{3cm}
\centering
\vspace*{-0.2\baselineskip}%
  \includegraphics[width=2.5cm]{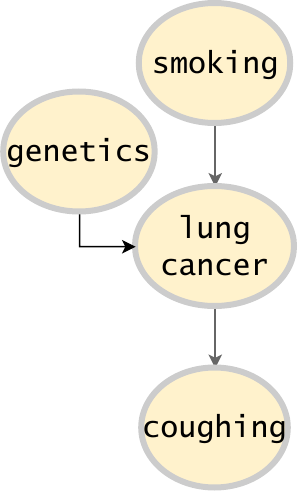}
  \caption{Example of a causal graph.}
  \label{fig:causalgraph}
\end{wrapfigure}

One of the fundamental tasks in various scientific disciplines is to find underlying causal relationships and eventually utilize them~\cite{glymour}. Causal discovery is a branch of causality study which estimates causal structures from observational data and generates a causal graph as a result. A causal graph, as illustrated in Fig.~\ref{fig:causalgraph}, is a directed graph modeling the causal relationships between observed variables; a node represents a variable and an edge represents a causal relationship.

Conventionally, causal discovery involves learning causal relations from observational data by measuring how changes in one variable are associated with changes in another variable, an approach referred to as \textit{covariance-based causal discovery}~\cite{kic2023causal}. 
Driven by the recent advancements in LLMs, recent work has explored the causal capabilities of LLMs using metadata (e.g., variable names) rather than their actual data values. In other words, the causal relation is queried in natural language directly to the LLMs. 
This paper focuses on the latter, and to differentiate with \textit{covariance-based causal discovery}, we refer to this approach as \textit{\textbf{knowledge-based causal discovery}}, following the definition of~\cite{kic2023causal}. 

Typically, such metadata-based causal reasoning is performed by Subject Matter Experts (SMEs) as they construct a causal graph, drawing from their expertise in domain-specific subjects and common sense~\cite{kic2023causal}, or based on literature surveys on subjects related to the variables. The advancement in LLMs has simplified this formerly challenging process, as LLMs are now capable of providing the knowledge that previously can only be provided by SMEs. Recent works~\cite{khetan-etal-2022-mimicause,tu2023causaldiscovery,willig2022foundation,zhang2023understanding} also show promising results, notably,~\cite{kic2023causal} explores causal capabilities of LLMs by experimenting on cause-effect pairs. Their finding suggests that LLM-based methods achieved state-of-the-art performance on several causal benchmarks. 
Similarly,~\cite{zhang2023understanding} investigated the causal capability of LLMs by analyzing their behavior given a certain causal question. However, in contrast to~\cite{kic2023causal}, their result suggests that LLMs currently lack the capability to offer satisfactory answers for discovering new knowledge. Meanwhile, a work by~\cite{khetan-etal-2022-mimicause} focused on investigating the LLMs' capability for causal association among events expressed in natural language. Thus, their study is more oriented towards extracting a causal diagram 
(e.g., a chain of events) from unstructured text instead of discovering new causal relations. 

In this paper, we investigate the capabilities of language models for knowledge-based causal discovery between variable pairs given a textual context from text sources.
Specifically, given a pair of variables $e_{1}$ and $e_{2}$, the task is to predict if a causal relation can be inferred between the variables. Therefore, similar to~\cite{khetan-etal-2022-mimicause}, 
our focus also lies in inferring causal relations from text rather than discovering new causal relations. In particular, we present ``\texttt{KG Structure as Prompt,}'' a novel approach for integrating structural information from a Knowledge Graph (KG) into \textit{prompt-based learning} with Small Language Models (SLMs). 
Prompt-based learning adapts LMs for specific tasks by incorporating prompts—task-specific instruction combined with the text input—to guide the models' output for the downstream tasks. Our approach enhances this method by incorporating additional information from KGs, leveraging the strengths of KGs in providing context and background knowledge. 
We opted for SLMs because a smaller model that can outperform larger models is more cost-effective and therefore preferable. We conduct experiments on three types of biomedical and an open-domain datasets, and further evaluate the performance of the proposed approach under three different architectures of language models.

To summarize, our main contributions are as follows:
\begin{enumerate}
    \item We present ``\texttt{KG Structure as Prompt}'', a novel approach for injecting structural information from KGs into prompt-based learning. 
     In experiments under few-shot settings, we demonstrate that our approach outperforms most of the no-KG baselines and achieves performance comparable to the conventional fine-tuning using a full dataset, even with limited samples. 
    
    \item %
    We show that our approach is effective with different types of language model architectures and knowledge graphs, showcasing its flexibility and adaptability across various language models and knowledge graphs. 
    
    \item We demonstrate the robust capabilities of SLMs: fused with prompt-based learning and an access to a knowledge graph, SLMs are able to surpass an LLM with much larger number of parameters.\footnote{\texttt{GPT-3.5-turbo} model~\cite{gpt35turbo} with ICL~\cite{GPT3NEURIPS2020_1457c0d6} prompting method} 
 
\end{enumerate}

\section{Background and Related Work}

\subsubsection{Small Language Models.}
\label{bg:slms}
Small Language Models (SLMs) refer to language models with fewer parameters, resulting in a reduced capacity to process text compared to larger-parameter LLMs. However, SLMs typically require less computation resources, 
making them faster to train and deploy, and maintaining them is generally more cost-effective. 
On the contrary, LLMs are trained on vast amounts of diverse data, thus have significantly more parameters and are capable of handling more complex language tasks than SLMs. 
Nevertheless, LLMs are expensive and difficult to train and deploy as they typically require more computational resource. For instance, GPT-3~\cite{GPT3NEURIPS2020_1457c0d6}, which consists of 175 billion parameters, is impractical to run on hardware with limited resources. 

In this work, we define SLMs as LMs with less than 1 billion parameters. We explore the causal capability of SLMs with different architectures: (1)~\underline{M}asked \underline{L}anguage \underline{M}odel (MLM) especially the encoder-only model, (2)~ \underline{C}ausal \underline{L}anguage \underline{M}odel (CLM) or decoder-only language model, and (3)~\underline{Seq}uence-\underline{to}-\underline{Seq}uence \underline{L}anguage \underline{M}odel (Seq2SeqLM) or encoder-decoder model. We provide an overview of each type of architecture below. 

MLMs, especially encoder-only models such as BERT~\cite{devlin-etal-2019-bert}, are a type of LM that utilizes encoder blocks within the transformer architecture and are trained to predict masked tokens based on the context provided by surrounding words. They excel in natural language understanding (NLU) tasks, e.g., text classification, as they are able to capture relationships between words in a text sequence. CLMs, such as GPT-3~\cite{GPT3NEURIPS2020_1457c0d6}, use the decoder blocks within the transformer architecture and are trained to generate text one token at a time, by conditioning each token on the preceding tokens in the sequence. 
Meanwhile, Seq2SeqLMs, such as T5~\cite{t5JMLR:v21:20-074}, consist of both encoder and decoder blocks. The encoder transforms the input sequence into vector representation, while the decoder produces the output based on the encoded vector. CLMs and Seq2SeqLMs generally work well for natural language generation (NLG) and NLU tasks such as translation and summarization, as they can produce coherent and grammatically accurate sentences.
We list our choice of language models in \S\ref{sec:slmchoice}.

\subsubsection{Prompt-based Learning \& Knowledge Injection.} 
\label{sec:crc}

Research on classifying causal relations from text has predominantly occurred within supervised settings, utilizing classical machine learning (ML) approaches~\cite{blanco-etal-2008-causal,Bui2010,CHANG2006662,Khoo2.10.3115/1075218.1075261,khoo1998,Mihăilă2014} or \textit{fine-tuning} pre-trained language models~\cite{Gu10.1093/database/baw042,khetan-etal-2022-mimicause,bioSu2022,medicause,yunicausal10.1145/3605098.3635894}. Classical ML techniques often require extensive feature engineering and have shown inferior performance compared to fine-tuning language models such as BERT~\cite{devlin-etal-2019-bert}. Therefore, we evaluate our method against fine-tuning methods as baselines. 

Meanwhile, \textit{\textbf{prompt-based learning}}, also known as \textit{prompt-tuning}, has recently emerged as a promising alternative to the conventional fine-tuning approach for a variety of Natural Language Processing (NLP) tasks~\cite{agrawal-etal-2022-large,pretrainprompt10.1145/3560815,schick-schutze-2021-exploiting,schick-schutze-2021-just}. 
Typically, a \textit{prompt} is composed of discrete text (\textit{hard} prompt); however, recent work has introduced \textit{soft prompt}, a continuous vector that can be optimized through backpropagation~\cite{ptlester-etal-2021-power,spli-liang-2021-prefix}. In the relation classification task, prompt-based learning often involves inserting a prompt template containing masked tokens into the input, essentially converting the task into \textit{masked language modeling} or \textit{text generation} problems~\cite{knowprompt10.1145/3485447.3511998,genpthan-etal-2022-generative,ptrHAN2022182}. This approach is particularly well-suited for few-shot or zero-shot scenarios, where only limited labeled data is available~\cite{gao-etal-2021-making,schick-schutze-2021-just}. 
This motivates us to investigate such prompt-based learning under few-shot settings, given the scarcity of datasets for our causal relation classification task. 

Other works explore \textbf{knowledge injection} for the prompt construction, for instance, KnowPrompt~\cite{knowprompt10.1145/3485447.3511998} injects latent knowledge contained in relation labels into prompt construction with learnable virtual words. KAPING~\cite{kapingbaek-etal-2023-knowledge} retrieves top-K similar triples of the target entities from Wikidata and further augments them as a prompt. KiPT~\cite{kiptli-etal-2022-kipt} uses WordNet to calculate semantic correlation between the input and manually constructed core concepts to construct the prompts. Our work differs from them since we focus on leveraging \textit{structural information} of knowledge graphs to construct the prompt (see \S\ref{sec:approach}).

\section{Task Formulation}
\label{sec:problemsetup}
In this work, we focus on \textit{pairwise} knowledge-based causal discovery: given a pair of entities $e_{1}$ and $e_{2}$, i.e., variable or node pairs such as \texttt{FGF6} and \texttt{prostate cancer}, the task is to predict if a causal relation can be inferred between the pair. We formulate the task as a \textit{binary classification task}, classifying the relation as \textit{causal} or \textit{non-causal}. We evaluate our approach on a dataset $\mathcal{D} =\{\mathcal{X}, \mathcal{Y}\}$, where $\mathcal{X}$ is a set of training instances and $\mathcal{Y}=\{causal, non\text{-}causal\}$ is a set of relation labels. Each instance $x \in \mathcal{X}$ consists of a token sequence $x=\{w_{1}, w_{2}, ... w_{|n|}\}$ and the spans of a marked variable pair, and is annotated with a label $y_{x} \in \mathcal{Y}$. 

\section{Approach}
\label{sec:approach}
\begin{figure}[tb]
  \centering
  \includegraphics[width=1\textwidth]{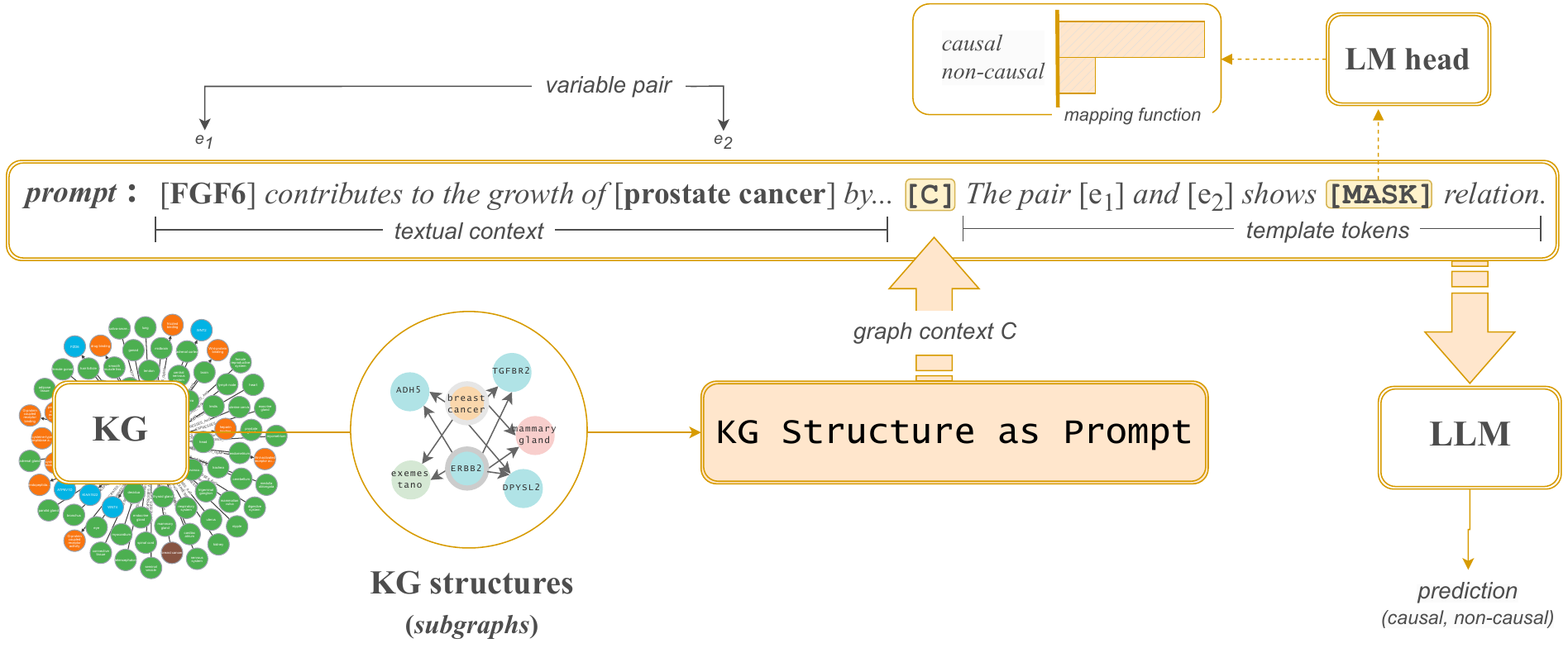}
\caption{Overall framework of our \texttt{KG Structure as Prompt} with prompt-based learning}
  \label{fig:arc}
\end{figure}

We illustrate our
proposed approach in Fig.~\ref{fig:arc}. 
First, we generate a \textbf{graph context}, which is derived from the structural information of a knowledge graph with our \texttt{KG Structure as Prompt} method. Next, we feed the generated graph context and the inputs, i.e., the \textbf{variable pair} and its \textbf{textual context},
into the SLMs to train a prompt-based learning model.

We elaborate our proposed approach in the following subsections. We start with preliminaries (\S\ref{prelim}), followed by the design of the \texttt{KG structure as Prompt} for generating the graph context (\S\ref{promptdesign}), and the incorporation of the generated graph context into the SLMs architecture with prompt-based learning (\S\ref{pblwithgc}).

\subsection{Preliminaries}
\label{prelim}
Formally, we define a directed graph $\mathcal{G} = (\mathcal{V}, \mathcal{E}$) where $\mathcal{V}$ is a set of vertices or nodes, and $\mathcal{E} \subseteq \mathcal{V} \times \mathcal{V}$ is a set of directed edges. A knowledge graph is a specific type of directed graph representing a network of entities and the relationships between them. Formally, we define a knowledge graph as a directed \textit{labeled} graph $\mathcal{KG} = (N, E, R, \mathcal{F})$ where $N$ is a set of nodes (entities), $E \subseteq N \times N$ is a set of edges (relations), $R$ is a set of relation labels, and $\mathcal{F}: E \to R$, is a function assigning edges to relation labels. For instance, assignment label $r$ to an edge $e=(x,y)$ can be viewed as a triple $(x, r, y)$, e.g., \texttt{(Tokyo, IsCapitalOf, Japan)}.

\subsection{Knowledge Graph Structure as Prompt}
\label{promptdesign}
In the field of Graph Neural Networks (GNNs), \cite{ye-etal-2024-language} explores whether LLMs can replace GNNs as the foundation model for graphs by using natural language to describe the geometric structure of the graph. 
Their results on several graph datasets surpass traditional GNN-based methods, showing the potential of LLMs as a new foundational model for graphs. Inspired by their success,
we similarly leverage the structural information of a specific type of graph, i.e., a knowledge graph, to infer causal relationships between variable pairs. We select knowledge graphs due to their rich structured information and their capability to express interconnected relationships. We call our approach \textbf{``Knowledge Graph Structure as Prompt''.}

\begin{figure}[h!]
  \centering
  \includegraphics[width=0.6\textwidth]{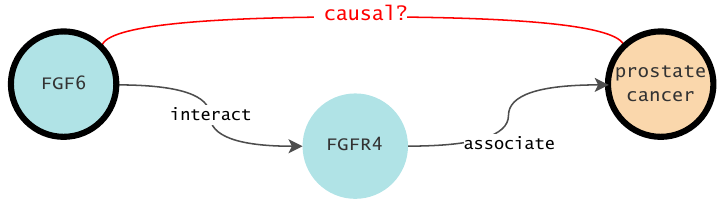  }
\caption{Illustration of inferring a causal relationship in KG.}
  \label{fig:hops}
\end{figure}

For instance, we may infer a causal relationship by looking at multi-hop relations between a node pair in a knowledge graph, as illustrated in Fig.~\ref{fig:hops}. 

In Fig.~\ref{fig:hops}, the node \texttt{FGF6} is \textit{indirectly} connected to the node \texttt{prostate cancer} within one hop via the node \texttt{FGFR4}. As verified by a human expert, there is indeed a causal relation between the nodes \texttt{FGF6} and \texttt{prostate cancer}. We argue that such graph structural information, in this example a \textit{path}, adds background knowledge on top of internal knowledge of LMs, effectively assisting LMs in inferring causal relation between the variable pair.
Specifically, we aim to use a natural language description of the structural information from the knowledge graph to be used as a prompt for prompt-based learning. We refer to such a description of knowledge graph structure as a \textit{\textbf{graph context}.}

In this work, we specifically examine three kinds of vital structural information of a KG to be used as the graph context, namely \textbf{(1) neighbor nodes, (2) common neighbor nodes, (3) \textit{metapath}}, described in detail as follows.

\subsubsection{(1) Neighbor Nodes ($\mathcal{NN}$).} The essence of GNNs lies in applying different aggregate functions to the graph structure, i.e., passing node features to \textit{neighboring nodes}, where each node aggregates the feature vectors of its neighbors to further update its feature vector. Thus, it is evident that the neighbor nodes are the most crucial feature within a graph. Inspired by that, we examine the neighboring nodes of the target node pairs to infer their causal relationship.

\begin{figure}[h!]
  \centering
  \includegraphics[width=0.75\textwidth]{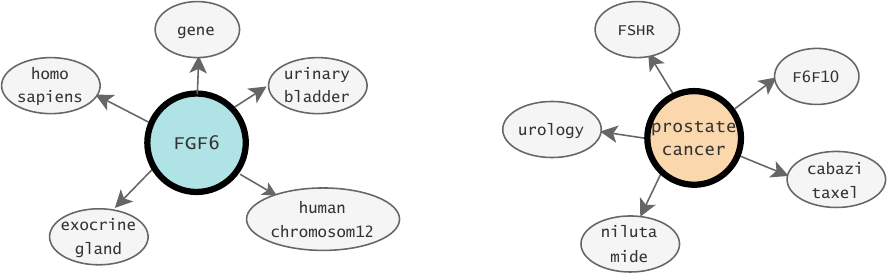  }
        \caption{Example of neighbor nodes for \texttt{FGF6} (\textit{left}) and \texttt{prostate cancer} (\textit{right}).}
  \label{fig:neigbors}
\end{figure}

Formally, a node $x$ is a \textbf{neighbor} of a node $y$ in a knowledge graph $\mathcal{KG} = (V, E)$ if there is an edge $\{x, y\} \in E$.
We provide an example of neighbor nodes from Wikidata~\cite{wikidata10.1145/2629489} in Fig.~\ref{fig:neigbors}.
According to the provided example, the node \texttt{prostate cancer} has \texttt{urology} as one of its neighbor nodes, while \texttt{FGF6} has \texttt{urinary bladder} as one of its neigbors. Thus, it is likely that a connection exists between the node pair (\texttt{FGF6, prostate cancer}) due to their respective neighboring nodes: \texttt{urinary bladde}r $\leftrightarrow$ \texttt{urology}.

For utilizing the neighbor nodes structure in the prompt, we describe it in natural language to form a graph context $\mathcal{C}$, which we formally denote as:
\begin{equation}
\label{eq:neighbors}
\mathcal{C}(x, V, E)=\{x\}\:\text{``is connected to''}\:\{[x_{2}]_{x_{2}\in V_2^x}\}
\end{equation}

We also create a variation of $\mathcal{C}$ where we include the edge description/relation labels $E$, formally denoted as follows:
\begin{equation}
\label{eq:neighbors2}
\mathcal{C}(x, V, E)=\{x\}\:\text{``has''}\:\{E_{x,x_{2}}\}\:\text{``relation with''}\:\{[x_{2}]_{x_{2}\in V_2^x}\}
\end{equation}

where $V_k^x$ represents the list of node $x$’s $k$-hop neighbor nodes and $E_{x,x_{2}}$ represents the relation or edge description between the node $x$ and its neighbor nodes. The additional template words such as \textit{``is connected to''} and \textit{``has relation with''} are optional, and can be replaced by any other fitting words. Then, with Eq.~\ref{eq:neighbors}, the generated graph context $C$ for the node \texttt{prostate cancer} in Fig.~\ref{fig:neigbors} is shown in the following Example~\ref{exnn}.

\begin{example}
\label{exnn}
    \texttt{prostate cancer} is connected to \textit{nilutamide, cabazitaxel, urology, FSHR, F6F10}
\end{example}

When including the relation labels (Eq.~\ref{eq:neighbors2}), the graph context would be:
\begin{example}
\label{exnn2}
    \texttt{prostate cancer} has \textit{drug or therapy used for treatment} relation with \textit{nilutamide and cabazitaxel}, has \textit{genetic association} with \textit{FSHR and F6F10}
\end{example}

\subsubsection{(2) Common Neighbor Nodes ($\mathcal{CNN}$).}
\begin{figure}[h!]
  \centering
  \includegraphics[width=0.38\textwidth]{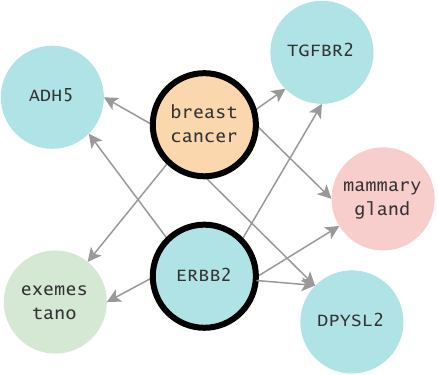  }
\caption{Example of common neighbor nodes for the pair \texttt{(ERBB2, breast cancer)}. Different colors of the nodes represent different node types.}
  \label{fig:cn}
\end{figure}

Unlike neighbor nodes, \textbf{common neighbor nodes} capture the idea that the more neighbors a pair of nodes $(x,y)$ shares, the more likely it is for the pair to be connected, i.e., there exists an edge $e = {x,y}$ between them. 
We argue that common neighbors between two nodes help infer their relationship, so we examine the common neighbors information between the node pair as graph context, as well. Fig.~\ref{fig:cn} shows an example of common neighbor nodes for the pair \texttt{(breast cancer}, \texttt{ERBB2)}, taken from Hetionet knowledge graph~\cite{hetionet10.7554/eLife.26726}. According to the provided example, the pair has in total 95 common neighbors, confirming a close relationship between them.

Formally, common neighbors between the nodes $x$ and $y$ can be defined as:
\begin{equation}
    \label{eq:cn}
    \mathcal{CN}\{x,y\} = N(x) \cap N(y)
\end{equation}

where $N(x)$ is the set of nodes adjacent to node $x$ (the neighbors of $x$), and $N(y)$ is the set of nodes adjacent to node $y$ (the neighbors of $y$). Subsequently, the graph context $\mathcal{C}$ for describing the common neighbors between the pairs $x$ and $y$ can be formed as follows:
\begin{equation}
\label{eq:cn2}
\mathcal{C}(x, \mathcal{CN}, y)=\text{``Common neighbor nodes of } \{x\}\:\text{and}\:\{y\}\:\text{are''}:\{[n]_{n\in \mathcal{CN}}\}
\end{equation}

where $\mathcal{CN}$ represents the list of common neighbor nodes of the pair $x$ and $y$ as defined in Eq.~\ref{eq:cn}. Again, the additional template words \textit{``Common neighbor nodes of...''} are optional and can be replaced by other words. Then, we can generate the graph context $\mathcal{C}$ including the common neighbor nodes information for the pair in Fig.~\ref{fig:cn}, as follows:
\begin{example}
\label{ex2}
    Common neighbor nodes of \texttt{breast cancer} and \texttt{ERBB2} are: \textit{ADH5, mammary gland, exemestane, TGFBR2, DPYSL2}
\end{example}

\subsubsection{(3) Metapath ($\mathcal{MP}$).}
\begin{figure}[h!]
  \centering
  \includegraphics[width=0.65\textwidth]{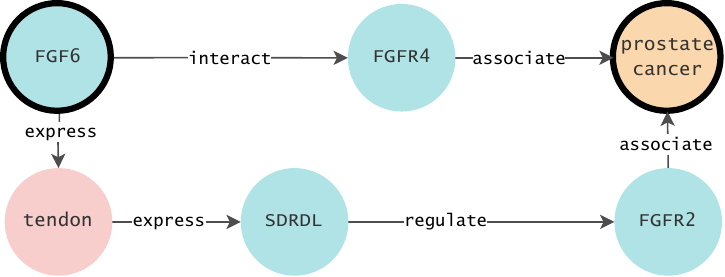  }
\caption{Example of metapaths for the pair \texttt{(FGF6, prostate cancer)}. Different colors of the nodes represent different node types.}
  \label{fig:metapath}
\end{figure}

\textit{Metapaths}, or \textit{meta-paths} are sequences of node types which define a walk from an origin node to a destination node~\cite{metapath10.14778/3402707.3402736}. The term ``metapath'' in this work is borrowed from the biomedical domain, referring to specific node type combinations thought to be informative~\cite{meta10.1093/bioinformatics/btad297}. Due to its importance in biomedical network analysis~\cite{wangmeta1,yaometa2}, we investigate the metapaths of two nodes for inferring their causal relationship. Moreover, causal relationships are frequently observed in the biomedical domain.
Fig.~\ref{fig:metapath} shows examples of metapaths of the pair \texttt{prostate cancer} and \texttt{FGF6}.

Formally, a metapath $\mathcal{MP}$ can be defined as a path $Z_{1}\xrightarrow{\mathit{R_{1}}}Z_{2}\xrightarrow{\mathit{R_{2}}} ...\xrightarrow{\mathit{R_{n}}}Z_{n+1}$ describing a relation $R$ between node types $Z$ and $Z_{n+1}$. The following examples illustrate metapaths of different path length $n$ from Fig.~\ref{fig:metapath}.
\begin{quote}
\text{$\mathcal{MP}$ = (\texttt{FGF6, FGFR4, prostate cancer})}, composed of node types \textit{\{gene, gene, disease\}}, with $n=3$
\end{quote}
\begin{quote}
\text{$\mathcal{MP}$ = (\texttt{FGF6, tendon, SDRDL, FGFR2, prostate cancer})}, composed of node types \textit{\{gene, anatomy, gene, gene, disease\}}, with $n=5$
\end{quote}

As explained by the example provided in Fig.~\ref{fig:hops}, we argue that an \textit{indirect path} between two nodes can be useful for inferring a causal relation between two pairs, even when the edge itself does not describe a causal relation. Thus, an \textit{indirect metapath}, or combination of meaningful node types could be \textit{latent} evidence of causality between a pair of variables. We describe the metapath structure from KG in natural language to form a graph context $\mathcal{C}$, as follows:
\begin{equation}
\label{eq:metapath}
\begin{aligned}
\mathcal{C}(x, y, \mathcal{MP}\{V,E\})&=\{x\}\:\text{``is connected to''}\:\{y\}\text{ ``via}\\
&\text{the following path: ''}\{v\} \: \{E_{v,v_{2}}\} \: \{v_{2}\}
\end{aligned}
\end{equation}

where $\mathcal{MP}\{V,E\}$ is a metapath $\mathcal{MP}$ with length $n$ containing a set of nodes $V(v_{1}, v_{2}, ... v_{|n|})$ and a set of edges $E(v_{1}, v_{1+n})$. Additional tokens \textit{``is connected to...''} are optional and can be replaced by other tokens. Then, the graph context $\mathcal{C}$ with metapath information for the example in Fig.~\ref{fig:metapath} would be:
\begin{example}
    \texttt{FGF6} is connected to \texttt{prostate cancer} via the following paths: \textit{FGF6 expressed in tendon, tendon expresses SQRDL, FGFR2 regulates SQRDL, FGFR2 associates with prostate cancer}
\end{example}

To prevent bias in prediction by the LMs, we avoid the \textit{direct} path.
For instance, for the pair $(x,y)$, we avoid the path $\mathcal{MP} = (x,y)$ and $\mathcal{MP} = (y,x)$, when such path exists in the KG.

\subsection{Prompt-based learning with \textit{graph context}} 
\label{pblwithgc}

As illustrated in the model architecture in Fig.~\ref{fig:arc}, we feed the \textbf{textual context} into SLMs together with the \textbf{graph context} generated with \texttt{KG Structure as Prompt}. We further design a prompt-based learning approach utilizing both contexts elaborated in this section. 
To get a clear distinction between conventional fine-tuning and our proposed prompt-based learning approach, we first provide a short overview of the conventional fine-tuning approach, as follows.

Given a pre-trained LM $\mathcal{L}$ to fine-tune on a dataset $\mathcal{D}$, the \textbf{conventional fine-tuning} method encodes the training sequence $x=\{w_{1}, w_{2}, ... w_{|n|}\}$ into the corresponding output hidden vectors of the LMs $h=\{h_{1}, h_{2}, ... h_{|n|}\}$. For MLMs such as BERT~\cite{devlin-etal-2019-bert}, the special token ``\texttt{[CLS]}'' is inserted at the beginning of the sequence, and this special token is used as the final sequence representation $h'$, since it is supposed to contain information from the whole sequence. A fully-connected layer and a softmax layer are further applied on top of this representation to calculate the probability distribution over the class set $\mathcal{Y}$, as follows.
\begin{equation}
    p=softmax(W_{f}h'+b_{f})
\end{equation}

\textbf{Prompt-based learning}, on the other hand, adapts the pre-trained LMs for the downstream task via priming on natural language \textit{\textbf{prompts}}$-$pieces of text that are combined with the input and fed to the LMs to produce an output for downstream tasks~\cite{agrawal-etal-2022-large}. Concretely, we first convert each input sequence $x$ with a template $\mathcal{T}$ to form a prompt $x'$: $\mathcal{T}:x\mapsto x'$. In addition, a mapping function $\mathcal{M}$ is used to map the downstream task class set $\mathcal{Y}$ to a set of label words $\mathcal{V}$ constituting all vocabularies of the LM $\mathcal{L}$, i.e., $\mathcal{M}:\mathcal{Y}\mapsto \mathcal{V}$. As in the pre-training of LMs, we further insert the special token ``\texttt{[MASK]}'' into $x'$ for $\mathcal{L}$ to fill with the label words $\mathcal{V}$. We provide an example of the prompt formulation below. \\

Given \text{$x$ =``Smoking causes cancer in adult male.''}, we set a template $\mathcal{T}$, e.g.,
\begin{quote}
\text{$\mathcal{T}$ = [$x$] ``It shows \text{\texttt{[MASK]}} relation.''} 
\end{quote}

Then, the prompt $x'$ would be: 
\begin{quote}
$x'$ = ``Smoking causes cancer in adult male. It shows \text{\texttt{[MASK]}} relation.''  
\end{quote}

We further feed the prompt $x'$ into $\mathcal{L}$ to obtain the hidden vector $h_{\text{\texttt{[MASK]}}}$ of \text{\texttt{[MASK]}}. Next, with the mapping function $\mathcal{M}$ connecting the class set $\mathcal{Y}$ and the label words, we formalize the probability distribution over $\mathcal{Y}$ at the masked position, i.e., $p(y|x) = p(\text{\texttt{[MASK]}} = \mathcal{M}(y)|x')$. Here, the mapping function can also be set manually e.g., $\mathcal{M}(true) = ``causal"$ and $\mathcal{M}(false) = ``non\text{-}causal"$. Note that depending on the task, dataset, and the prompt design, the class labels themselves can be used directly without any mapping function $\mathcal{M}$.

In this study, our prompt-based learning combines the input sequence $x$ with the graph context $\mathcal{C}$ into the prompt $x'$, 
as illustrated in Fig.~\ref{fig:arc}. Specifically, we formulate the prompt $x'$ to include the following elements: 
\begin{enumerate}
    \item [(1)] textual context: input sequence $x$ containing the pair,
    \item [(2)] graph context $\mathcal{C}$: context generated from KG structures as described in \S\ref{promptdesign},
    \item [(3)] target pair: pair $e_{1}$ and $e_{2}$ as the target, e.g., \texttt{(FGF6, prostate cancer)},
    \item [(4)] \texttt{[MASK]} token,
    \item [(5)] (\textit{optional}) template tokens.
\end{enumerate}

Subsequently, our final prompt $x'$ as the input to the LM for the pair $e_{1}$ and $e_{2}$ can be formally defined as: 
\begin{equation}
\label{eq:mlm}
x' = [x] \: [\mathcal{C}] \: \text{The pair}\:[e_{1}]\:\text{and}\:[e_{1}]\:\text{shows a}\:\text{\texttt{[MASK]}} \:\text{relation.} 
\end{equation}

In this study, we select three SLMs, one for each of the three architectures: MLM, CLM, Seq2SeqLM. Since each type of SLMs is trained differently, we design the prompt $x'$ differently across each type of SLMs. For instance, the prompt $x'$ in Eq.~\ref{eq:mlm}, which is a \textit{cloze-style task} prompt, suits the MLM architecture, since this model is trained to be able to see the preceding and succeeding words in texts. As for CLM and Seq2SeqLM, we cast the task as a \textit{generation}-type, with prompt $x'$ such as:
\begin{equation}
\label{eq:clm}
x' = [x] \: [\mathcal{C}] \: \text{The pair }[e_{1}]\text{ and }[e_{2}]\text{ shows a causal relation: }\text{\texttt{[MASK]}}.
\end{equation}

As mentioned earlier, the design of the mapping function $\mathcal{M}$ to map the output into the downstream task labels varies depending on the task, dataset, and the prompt design. For instance, with the prompt $x'$ as in Eq.~\ref{eq:mlm}, we can directly use the class labels set $\mathcal{Y}=\{causal, non\text{-}causal\}$ without any mapping function. Meanwhile, for prompt $x'$ in Eq.~\ref{eq:clm}, we manually define a mapping function, e.g., $\mathcal{M}(causal) = ``true"$ and $\mathcal{M}(non\text{-}causal) = ``false"$. Note that the template ``\textit{The pair shows...}'' is optional and can be replaced with other text. \\

\section{Evaluation}
\subsubsection{Experiment Settings.} 
We evaluate the proposed approach under few-shot settings, using $k=16$ training samples across all experiments. Precision (P), Recall (R), and F1-score (F1) metrics are employed to evaluate the performance.
Since fine-tuning on low resources often suffers from instability and results may change given a different split of data~\cite{schick-schutze-2021-just}, we apply 5-fold cross-validation and the metric scores are averaged. We restrict the number of contents from the KG structures to be included in the prompt since the length of the prompt for SLMs is limited, and we restrict the number of hops when querying the KG, as well. 
We experimented with different settings and reported the best performing models. Additional technical details are provided online
as supplementary materials.\
\subsubsection{Datasets.}
\label{sec:data}

The evaluation datasets are summarized in Table~\ref{tab:datastat}. Causality is often observed in the biomedical domain, thus we primarily evaluate our approach within this field, supplemented by an open-domain dataset. 
Each instance in the dataset comprises textual context where a variable pair co-occurs in a text (see Example~\ref{ex:data} \& \ref{ex:data2}), 
and is annotated by human experts to determine if there is a causal relation between the variables. 

\begin{example}
\label{ex:data}
\textit{\textbf{FGF6} contributes to the growth of \textbf{prostate cancer} by activating...}
\end{example}

\begin{example}
\label{ex:data2}
\textit{The deadly train \textbf{crash} was caused by a terrorist \textbf{attack}.}
\end{example}

\begin{table}
    \centering
    \caption{Dataset sizes and types.}
    \label{tab:datastat}
    \vspace{-1mm}
    \begin{tabular}{lcrc}
    \toprule
      dataset & domain & total instances & description\\
    \midrule
    GENEC (ours) & biomedical & 789 & gene-gene causality\\
    DDI~\cite{ddiHERREROZAZO2013914}  &  biomedical &33,508 & drug-drug causality\\
    COMAGC~\cite{comaLee2013com} &  biomedical &820 & gene-disease causality\\
    SEMEVAL-2010 Task 8~\cite{semhendrickx-etal-2010-semeval}  &  open-domain & 10,717 & general domain causality\\
    \bottomrule
    \end{tabular}
\end{table}

\subsubsection{\textbf{Choice of SLMs.}}
\label{sec:slmchoice} 
In this work, we define SLMs as LMs with \textbf{less than 1 billion }of total parameters. We experimented with SLMs with three different architectures, as follows.
\begin{itemize}
    \item [(a)] MLM: \texttt{roberta}~\cite{robertaDBLP:journals/corr/abs-1907-11692} model adapted to the biomedical domain, with 125 million parameters (\texttt{biomed-roberta-base-125m}~\cite{biorobertagururangan-etal-2020-dont}),
    \item [(b)] CLM: \texttt{bloomz-560m}~\cite{bloomzmuennighoff2023crosslingual} with 560 million parameters,
    \item [(c)] Seq2SeqLM: \texttt{T5-base-220m}~\cite{t5JMLR:v21:20-074} model with 220 million parameters
\end{itemize}

\subsubsection{Choice of KGs.}
\label{sec:kg}
We selected the following two KGs for the experiments: 
\begin{itemize}
    \item [(a)] Wikidata~\cite{wikidata10.1145/2629489}, as a representation of the general-domain KG,
    \item [(b)] Hetionet~\cite{hetionet10.7554/eLife.26726}, a domain-specific KG assembled from 29 different databases, covering genes, compounds, and diseases.
\end{itemize}
We selected Wikidata for its broad coverage of numerous subjects and topics. As a comparison, we selected biomedical-KG Hetionet since we primarily evaluate our approach on the datasets from this particular domain.

\subsubsection{Model Comparison.}
We compare the following models: Models (1) to (4) represent the models trained without the graph context, i.e., the baselines, while models marked with ``PBL'' (model 5 to 7) are our proposed prompt-based learning method injected with graph context information from KGs. 

\begin{itemize}
    \item [(1)] ICL: \underline{I}n-\underline{C}ontext \underline{L}earning refers to a prompting method where few demonstrations of the task are provided to the LLMs as part of the prompt~\cite{GPT3NEURIPS2020_1457c0d6}. For this method, we selected \texttt{GPT-3.5-turbo-instruct} model by OpenAI, and provided $k=16$ samples as demonstrations to query the model.
    \item [(2)] FT$_{\text{full}}$: Conventional \underline{F}ine-\underline{T}uning models trained using the full datasets.
    \item [(3)] FT$_{\text{few-shot}}$: Conventional \underline{F}ine-\underline{T}uning models under few-shot $k=16$ setting.
    \item [(4)] PT$_{\text{few-shot}}$: Original \underline{P}rompt \underline{T}uning~\cite{ptlester-etal-2021-power} as a baseline, which is essentially prompt-based learning without any graph context. 
    \item [(5)] PBL$_{\text{$\mathcal{NN}$-Wiki-few-shot}}$: Our proposed \underline{P}rompt-\underline{b}ased \underline{L}earning $+$ \texttt{KG Structure as Prompt} using the neighbor nodes $\mathcal{NN}$ structure from Wikidata.
    \item [(6)] PBL$_{\text{$\mathcal{CNN}$-Het-few-shot}}$: Our proposed \underline{P}rompt-\underline{b}ased \underline{L}earning $+$ \texttt{KG Structure as Prompt} using the common neighbor nodes $\mathcal{CNN}$ structure from Hetionet.
    \item [(7)] PBL$_{\text{$\mathcal{MP}$-Het-few-shot}}$: Our proposed \underline{P}rompt-\underline{b}ased \underline{L}earning $+$ \texttt{KG Structure as Prompt} using the metapaths $\mathcal{MP}$ structure from Hetionet.
\end{itemize}

Some variants of the proposed approach, such as PBL$_{\text{$\mathcal{MP}$-Wiki-few-shot}}$, were omitted due to the computational expense of multi-hop querying in Wikidata, which often results in no usable metapaths for many pairs. In addition, to focus the discussion on the more effective models, we have omitted the results of less effective models, such as PBL$_{\text{$\mathcal{NN}$-Het-few-shot}}$. We provide the more complete results, including zero-shot prompting and classical ML experiment results, online as supplementary materials.

\section{Results and Discussion}
\label{sec:resultdis}
Table~\ref{tab:result} \&~\ref{tab:result_sem} summarize the results. We report the averaged Precision (P), Recall (R), and F1 scores, including the standard deviation values of the F1 scores over the 5-folds cross-validation. We provide a summary of the primary findings (\S\ref{finding}), followed by analysis and discussion of the results (\S\ref{disc}).

\begin{table}[h!]
\centering
\caption{Evaluation results on biomedical-domain datasets. Values in parenthesis are the standard deviations of F1 scores over 5-cv test folds. 
$_{\mathcal{NN}}$, $_{\mathcal{CNN}}$, $_{\mathcal{MP}}$ indicate the KG structures used as \textit{graph context}: \textit{neighbors nodes}, \textit{common neighbors nodes}, and \textit{metapath}, respectively. \textbf{bold}: highest F1 scores per LMs architecture and per dataset, \underline{underline}: F1 scores of the highest-performed models per dataset.} 
\vspace{-1mm}
\label{tab:result}
\begin{tabular}{lccccccccc}
\toprule
& \multicolumn{3}{c}{COMAGC} & \multicolumn{3}{c}{GENEC} & \multicolumn{3}{c}{DDI}  \\ \midrule
 & P       & R       & F1     & P      & R      & F1     & P      & R      & F1 \\ \midrule
(baseline) ICL$_{\text{(GPT-3.5-turbo)}}$ & 64.1 & 67.7 & $65.5_{(.18)}$    
& 55.4 & 77.2 & $62.6_{(.16)}$         
& 53.2 & 99.0 & $68.9_{(.08)}$ \\
\toprule
\multicolumn{10}{c}{*MLM architecture (\texttt{biomed-roberta-base-125m})*} \\
\midrule
FT$_{\text{full}}$ & 90.0 & 86.8 & $88.2_{(.02)}$ & 61.0 & 61.0 & $61.5_{(.03)}$ & 84.0 & 98.9 & $90.7_{(.04)}$ \\
(baseline) FT$_{\text{few-shot}}$ & 87.0 & 71.0 & $76.8_{(.02)}$ & 52.0 & 52.0 & $52.0_{(.02)}$ & 64.9 & 87.0 & $73.9_{(.07)}$ \\
(baseline) PT$_{\text{few-shot}}$ & 87.0 & 77.2 & $80.6_{(.03)}$ & 58.0 & 58.0 & $58.0_{(.02)}$& 69.6 & 83.0 & $75.4_{(.03)}$ \\
(ours) PBL$_{\text{$\mathcal{NN}$-Wiki-few-shot}}$   & 82.0 & 85.0 & $83.5_{(.02)}$  & 63.0 & 63.0 & \textbf{63.0}$_{(.03)}$& 70.2 & 85.0 & $\textbf{\underline{76.6}}_{(.03)}$ \\
(ours) PBL$_{\text{$\mathcal{CNN}$-Het-few-shot}}$ & 77.4 & 89.5 & \textbf{$82.7_{(.05)}$} & 54.5 & 54.4 & $54.4_{(.02)}$& 65.1 & 89.9 & $74.6_{(.05)}$ \\
(ours) PBL$_{\text{$\mathcal{MP}$-Het-few-shot}}$ & 80.0 & 89.3 & $\textbf{\underline{83.9}}_{(.03)}$ & 60.5 & 60.5 & $60.5_{(.02)}$& 67.0 & 87.6 & $75.4_{(.02)}$ \\                       
\toprule
\multicolumn{10}{c}{*CLM architecture (\texttt{bloomz-560m})*} \\
\midrule
FT$_{\text{full}}$ & 58.0 & 91.8 & $71.3_{(.07)}$ & 53.8 & 73.0 & $60.7_{(.07)}$ & 83.0 & 91.4 & $86.7_{(.06)}$ \\
(baseline) FT$_{\text{few-shot}}$   & 73.4 & 85.0 & $77.6_{(.04)}$ & 48.2 & 86.0 & $61.7_{(.01)}$ & 72.0 & 61.5 & $66.1_{(.04)}$ \\
(baseline) PT$_{\text{few-shot}}$ & 64.1 & 95.0 & $76.3_{(.01)}$ & 51.3 & 97.0 & $67.6_{(.01)}$  & 60.2 & 92.0 & $72.7_{(.02)}$ \\
(ours) PBL$_{\text{$\mathcal{NN}$-Wiki-few-shot}}$   & 71.7 & 90.9 & $79.0_{(.02)}$  & 54.1 & 90.0 & $67.4_{(.01)}$ & 63.3 & 88.0 & $72.6_{(.02)}$  \\
(ours) PBL$_{\text{$\mathcal{CNN}$-Het-few-shot}}$ & 65.7 & 94.7 & \textbf{$77.2_{(.05)}$} & 52.8 & 100 & \textbf{69.1}$_{(.00)}$& 60.4 & 93.9 & $72.8_{(.05)}$ \\
(ours) PBL$_{\text{$\mathcal{MP}$-Het-few-shot}}$ & 76.9 & 89.0 & $\textbf{81.8}_{(.04)}$  & 52.8 & 91.0 & $68.5_{(.03)}$ & 65.5 & 90.8 & \textbf{74.8}$_{(.05)}$ \\
\toprule
\multicolumn{10}{c}{*Seq2SeqLM architecture (\texttt{T5-base-220m})*} \\
\midrule
FT$_{\text{full}}$ & 88.5 & 78.0 & $82.6_{(.07)}$  
& 60.6 & 43.0 & $50.0_{(.11)}$  
& 96.3 & 83.0 & $88.8_{(.04)}$ \\
(baseline) FT$_{\text{few-shot}}$   & 68.8 & 81.0 & $72.7_{(.06)}$ 
& 48.3 & 67.0 & $55.5_{(.01)}$  
& 55.6 & 94.0 & $69.6_{(.04)}$  \\
(baseline) PT$_{\text{few-shot}}$ & 64.5 & 91.0 & $75.0_{(.05)}$  
& 63.0 & 83.0 & $69.3_{(.06)}$ 
& 63.3 & 82.0 & $70.3_{(.04)}$  \\
(ours) PBL$_{\text{$\mathcal{NN}$-Wiki-few-shot}}$   & 65.3 & 90.0 & $74.7_{(.01)}$  
& 66.5 & 76.0 & $70.3_{(.02)}$ 
& 57.5 & 95.0 & $70.9_{(.03)}$ \\
(ours) PBL$_{\text{$\mathcal{CNN}$-Het-few-shot}}$ & 65.4 & 86.3 & \textbf{$73.4_{(.02)}$} 
& 56.0 & 95.8 & $\textbf{\underline{70.6}}_{(.01)}$
& 56.2 & 96.0 & $70.6_{(.02)}$ \\
(ours) PBL$_{\text{$\mathcal{MP}$-Het-few-shot}}$ & 73.3 & 82.0 & \textbf{75.6}$_{(.03)}$  
& 56.5 & 91.0 & $69.7_{(.02)}$
& 60.6 & 92.9 & \textbf{72.9}$_{(.03)}$  \\

\bottomrule
\end{tabular}
\end{table}

\begin{table}[h!]
\centering
\caption{Evaluation results for open-domain dataset: SEMEVAL-2010 Task 8} 
\vspace{-1mm}
\label{tab:result_sem}
\begin{tabular}{lccccccccc}
\toprule

 & P       & R       & F1     & P      & R      & F1     & P      & R      & F1 \\
 \midrule
(baseline) ICL$_{\text{GPT-3.5-turbo}}$   
& 52.2 & 97.2 & $67.5_{(.07)}$ \\
\midrule
& \multicolumn{3}{c}{MLM} & \multicolumn{3}{c}{CLM} & \multicolumn{3}{c}{Seq2SeqLM}  \\
\midrule
FT$_{\text{full}}$ 
& 94.8 & 88.7 & $91.6_{(.02)}$  
& 73.0 & 74.0 & $71.3_{(.07)}$ 
& 82.8 & 77.0 & $79.4_{(.02)}$ \\

(baseline) FT$_{\text{few-shot}}$   
& 57.2 & 83.0 & $67.2_{(.04)}$ 
& 50.2 & 99.0 & $66.6_{(.01)}$  
& 52.1 & 90.9 & $61.1_{(.04)}$  \\

(baseline) PT$_{\text{few-shot}}$ 
& 51.9 & 94.0 & $66.9_{(.00)}$  
& 51.7 & 98.0 & $\textbf{67.6}_{(.02)}$ 
& 52.5 & 93.0 & $66.7_{(.02)}$  \\

(ours) PBL$_{\text{$\mathcal{NN}$-Wiki-few-shot}}$   
& 55.7 & 93.0 & $\textbf{\underline{69.7}}_{(.01)}$  
& 53.4 & 91.2 & $67.4_{(.00)}$ 
& 53.7 & 93.0 & $\textbf{67.9}_{(.01)}$ \\
\bottomrule
\end{tabular}
\end{table}

\subsection{Primary Findings} 
\label{finding}
We listed the summary of the main results from Table~\ref{tab:result} below.
\begin{itemize}
    \item [(a)] Our proposed approach outperforms the no-graph context baseline models, in most of the experiments across different dataset domains: up to $15.1$ points of improvement of F1 scores on biomedical datasets ($55.5\xrightarrow{} 70.6$, GENEC dataset) and $6.8$ points of improvement on open-domain dataset ($61.1\xrightarrow{} 67.9$, SEMEVAL dataset). 
    \item [(b)] Under few-shot settings with $k=16$ training samples, our proposed approach generally achieves the second-best performance compared to FT$_{\text{full}}$ model, which is the conventional fine-tuning models trained with full datasets.
    Few even surpassed them, such as PBL$_{\text{$\mathcal{NN}$-Wiki-few-shot}}$ model with MLM architecture (63.0 vs. 61.5) and PBL$_{\text{$\mathcal{CNN}$-Het-few-shot}}$ model with CLM architecture, on GENEC dataset (69.1 vs. 60.7).
    \item [(c)] Our models based on SLMs with less that 1 billion parameters surpassed the ICL prompting method on much larger model across all datasets in most experiments, underlining the importance of KGs to support the constrained internal knowledge of smaller LMs. 
\end{itemize}

\subsection{Analysis and Discussion}
\label{disc}
\subsubsection{$\mathcal{NN}$ vs. $\mathcal{CNN}$ vs. $\mathcal{MP}$ } In our experiments, the KG structure \textit{metapath} $\mathcal{MP}$ contributed the most to the top-performing models, while the neighbors nodes $\mathcal{NN}$ and common neighbors nodes $\mathcal{CNN}$ roughly exhibited comparable performance across models and datasets. The effectiveness of $\mathcal{MP}$ likely depends on the hop count between entity pairs in the dataset, i.e., the hop count is relatively high (2.8) for the COMACG dataset, where $\mathcal{MP}$ gave the best performance. Conversely, for the GENEC dataset, where the average $\mathcal{MP}$ hop is low (1.8), $\mathcal{CNN}$ and $\mathcal{NN}$ outperformed $\mathcal{MP}$. To train a robust model that is able to generalize well given any KG structure, we opted to not optimize the content selection of the KG structures in the current experiments. For instance, when there are more than $m$ metapaths for a pair, we \textit{randomly} select $m$ of them, $m$ being a \textit{hyperparameter} of the number of metapaths to be included as prompt. In spite of that, our proposed approach achieved a relatively satisfactory performance, suggesting that rather than the content of the structure, the \textbf{type of the structural information}, i.e., $\mathcal{NN}$ vs. $\mathcal{CNN}$ vs. $\mathcal{MP}$, is arguably more important based on the experiments.
\subsubsection{MLM vs. CLM vs. Seq2SeqLM.} For classification tasks, language models trained with MLM architecture are often preferred. This preference comes from the fact that MLMs are trained to consider both preceding and succeeding words, a crucial aspect for accurately predicting the correct class in a classification model. In line with this, the top-performing models trained on both full and few-shot datasets are based on the MLM architecture. The second best-performing models using full dataset are based on the Seq2SeqLM architecture, followed by those based on the CLM architecture. This is most likely because, similar to MLMs, Seq2SeqLMs also include encoder blocks and are trained to recognize the surrounding words~\cite{t5JMLR:v21:20-074}. However, this trend slightly differs in experiments under few-shot settings, as the models based on CLM architecture outperformed those based on Seq2SeqLM architecture.
Thus, selecting an appropriate architecture, specifically how the LMs are trained, is crucial when adapting the LMs for downstream tasks. As demonstrated by the outcomes of our experiments, LMs trained with the MLM architecture are generally more suitable for classification tasks than those with Seq2SeqLM and CLM architectures.

\subsubsection{Wikidata vs. Hetionet.}  
In the biomedical domain, the proposed approach injected with structural information from Hetionet demonstrates better performance in most experiments. This is expected considering the domain-specific nature of the dataset. Nevertheless, both Wikidata and Hetionet performed relatively well; the top-performing models for COMAGC and GENEC datasets are attained with Hetionet, while for DDI dataset are achieved with Wikidata. We also achieved 6.8 points of F1 score improvement on SEMEVAL dataset with Wikidata. This suggests that the proposed approach is rather flexible regarding the choice of KGs.

\subsubsection{SLMs vs. LLMs.} We selected OpenAI's \texttt{GPT-3.5-turbo-instruct}~\cite{gpt35turbo} as a representative of larger parameter-LLMs. However, OpenAI does not provide technical details such as the numbers of parameters; except the context windows which is 4,096 tokens in size for this model~\cite{gpt35turbo}. This is much larger than our choice of SLMs with a maximum token length ranging from 128 to 512. To summary, the results demonstrate that the SLMs outperformed this model across all datasets in most experiment. This further shows the potential of SLMs: combined with prompt-based learning and access to KGs, the proposed approach outperforms LLMs with considerably larger size and parameters, with minimal training effort (few-shot). Note that we also provided $k=16$ training samples as task demonstration to query the GPT model for a more fair comparison with the experiments under few-shot settings. 

Typically, SLMs are trained on significantly less data compared to LLMs, which leads to reduced capacity and inferior performance in downstream tasks. Therefore, the graph context derived from the structural information of KG by our proposed \texttt{KG Structure as Prompt} approach effectively serves as an additional \textit{evidence of causality}; in other words, it assists the SLMs to rely not only on their constrained internal knowledge, but also by enhancing their capacity through denser information sourced from the KGs. 

\section{Conclusion}
In this paper, we presented \textbf{``\texttt{KG Structure as Prompt}''}, a novel approach for integrating structural information from KGs into prompt-based learning, to further enhance the capability of Small Language Models (SLMs). We evaluated our approach on knowledge-based causal discovery tasks. Extensive experiments under few-shot settings on biomedical and open-domain datasets highlight the effectiveness of the proposed approach, as it outperformed most of the no-KG baselines, including the conventional fine-tuning method with a full dataset. We also demonstrated the robust capabilities of SLMs: in combination with prompt-based learning and KGs, SLMs are able to surpass a language model with larger parameters. Our proposed approach has proven to be effective with different types of LMs architectures and KGs, as well, showing its flexibility and adaptability across various LMs and KGs. 

Our work has been centered on discovering causal relationships between pairs of variables. In future work, we aim to tackle more complex scenarios by developing methods to analyze causal graphs with multiple interconnected variables, which will offer a deeper understanding of causalities. %
\paragraph*{Supplemental Material Statement:} 
\label{sup}
Datasets, source code, and other details are available online at \url{https://github.com/littleflow3r/kg-structure-as-prompt}

\bibliographystyle{splncs04}
\bibliography{ref-all} %
\end{document}


\title{(APPENDIX) \\
Knowledge Graph Structure as Prompt: Improving Small Language Models Capabilities for Knowledge-based Causal Discovery}

\titlerunning{\texttt{KG Structure as Prompt}: SLMs for Knowledge-based Causal Discovery}

\appendix
\noindent \textit{The subsequent sections constitutes the appendix.}

\section{Additional results: Classical ML models}
For comprehensiveness, we trained three classical ML models—SVM, Logistic Regression (LR), and Random Forest (RF)—using simple n-gram/tf-idf features on our causal relation classification task, for both the biomedical and general domain datasets. Please note that we trained the classical ML models using the \textbf{full training examples} rather than the 16 examples used in our few-shot experiments. The result is summarized in Table~\ref{tab:resultml}. 
\begin{table}[h!]
\centering
\caption{Evaluation results with classical ML models} 
\vspace{-1mm}
\label{tab:resultml}
\begin{tabular}{lccccccccc}
\toprule
& \multicolumn{3}{c}{\texttt{SVM}} & \multicolumn{3}{c}{\texttt{LR}} & \multicolumn{3}{c}{\texttt{RF}}  \\
\midrule
 & P       & R       & F1     & P      & R      & F1     & P      & R      & F1 \\
 \midrule
COMAGC
& 71.4 & 25.0 & 37.0
& 100 & 10.0 & 18.1
& 100 & 15.0 & 26.1 \\

GENE   
& 60.0 & 15.0 & 24.0  
& 0.0 & 0.0 & 0.0 
& 50.0 & 50.0 & 9.00 \\

DDI 
& 59.3 & 95.0 & 73.0 
& 59.3 & 95.0 & 73.0 
& 60.0 & 90.0 & 72.0 \\

SEMEVAL
& 75.0 & 30.0 & 42.8  
& 63.6 & 35.0 & 45.1 
& 52.1 & 60.0 & 55.8 \\
\bottomrule
\end{tabular}
\end{table}

Overall, the classical ML models performed significantly poorer compared to Transformer/LM-based models (Table 2 and 3 in the main paper), even when trained on the full training data. However, it performed relatively well on the DDI datasets, achieving an F1 score of 73.0 with SVM and LR models, which outperformed the ICL model using GPT-3.5 (F1=68.9). We believe this discrepancy is due to the relatively small size of the other datasets compared to the DDI datasets, which proved that classical ML models tend to rely on larger amounts of data for effective training.

Another challenge on using classical ML approach is that it often requires complex feature engineering, while in deep learning-based approach such as Transformer-based model, the high-dimensional features are learned automatically by the "deep"-layered model, often resulting in better accuracy. 

\section{Additional results: zero-shot prompting}
We included the results of zero-shot prompting, summarized in Table~\ref{tab:result0}. As in few-shot prompting (i.e., the ICL model), we used the~\texttt{gpt-3.5-turbo-instruct} LLM for zero-shot prompting, as well. To summary the results, the zero-shot baselines have shown inferior performance compared to the other models (few-shot and full fine-tuning).
\begin{table}[h!]
\centering
\caption{Evaluation results: zero-shot prompting} 
\vspace{-1mm}
\label{tab:result0}
\begin{tabular}{lccc}
\toprule
 & P       & R       & F1 \\
 \midrule
COMAGC
& 51.7 & 75.0 & 61.2 \\

GENE   
& 50.0 & 15.0 & 23.0 \\

DDI 
& 40.0 & 20.0 & 26.6 \\

SEMEVAL
& 53.3 & 80.0 & 64.0 \\
\bottomrule
\end{tabular}
\end{table}

\section{Baseline (\texttt{GPT-3.5-turbo-instruct}) hyperparameters}

In this work, we used OpenAI\footnote{https://platform.openai.com/} API with \texttt{gpt-3.5-turbo-instruct} engine. In general we assume only query access to the LLMs (i.e., no gradients, no log probabilities). Table~\ref{tab:llmparam} summarizes the hyperparameter values for the ICL model experiment with \texttt{gpt-3.5-turbo-instruct}. 

\begin{table}[h!]
\centering
  \caption{Hyperparameter values.}
  \label{tab:llmparam}
  \begin{tabular}{lc}
    \toprule
    \textit{hyperparameter} & value \\
    \midrule    
    \texttt{temperature} & 0.7 \\
    \texttt{max token} & $100\sim400$ \\
    \texttt{top p} & 1 \\
    \texttt{frequency penalty} & 0 \\
    \texttt{presence penalty} & 0 \\
    \bottomrule
  \end{tabular}
\end{table}

\section{MLM hyperparameters settings}
We used the \texttt{biomed\_roberta\_base}\footnote{\url{https://huggingface.co/allenai/biomed_roberta_base}} for all biomedical datasets (COMAGC, GENEC, and DDI), and \texttt{roberta-base}\footnote{\url{https://huggingface.co/FacebookAI/roberta-base}} for the SEMEVAL dataset. We used both models from the \texttt{huggingface}~\cite{huggingface-wolf-etal-2020-transformers} models library. All models are implemented in Python with \texttt{Pytorch}\footnote{\url{https://pytorch.org/}} and \texttt{Transfomer}\footnote{\url{https://huggingface.co/docs/transformers/en/index}} library. The random seed of 203 is set for all experiments. 
Table~\ref{tab:bertparam} summarizes the hyperparameter values for fine-tuning the model experiments. 

\begin{table}[h!]
\centering
  \caption{MLM hyperparameter values.}
  \label{tab:bertparam}
  \begin{tabular}{lcccc}
    \toprule
     \textit{hyperparameter} & COMAGC & GENEC & DDI & SEMEVAL \\
    \midrule
    \texttt{batch size} & 8 & 8 & 8 & 8 \\
    \texttt{max length} & 256 & 256 & 256 & 256 \\
    \texttt{optimizer} & \texttt{Adam} & \texttt{Adam} & \texttt{Adam} & \texttt{Adam} \\
    \texttt{lr} & 3e-5 & 3e-5 & 3e-5 & 3e-5 \\
    \texttt{gradient acc. steps} & 4 & 4 & 4 & 4 \\
    \texttt{adam eps.} & 1e-06  & 1e-06  & 1e-06  & 1e-06  \\
    \texttt{warmup proportion} & 0.06 & 0.06 & 0.06 & 0.06 \\
    \texttt{weight decay} & 0.01  & 0.01  & 0.01  & 0.01  \\
    \texttt{max grad norm} & 1  & 1  & 1  & 1  \\
    \texttt{epoch} & 30 & 30 & 30 & 20 \\
    \bottomrule
  \end{tabular}
\end{table}

\section{CLM/decoder-only hyperparameters settings}
In all evaluation experiments, the \texttt{bloomz-560m}\footnote{\url{https://huggingface.co/bigscience/bloomz-560m}}  model from the \texttt{huggingface} models library were used for all datasets. All models are implemented in Python with \texttt{Pytorch}\footnote{\url{https://pytorch.org/}} and \texttt{Transfomer}\footnote{\url{https://huggingface.co/docs/transformers/en/index}} library. The random seed of 203 is set for all experiments. Table~\ref{tab:bloomparam} summarizes the hyperparameter values for fine-tuning the \texttt{bloomz-560m} model experiments. 

\begin{table}[h!]
\centering
  \caption{\texttt{bloomz-560m} hyperparameter values.}
  \label{tab:bloomparam}
  \begin{tabular}{lcccc}
    \toprule
     \textit{hyperparameter}     & COMAGC & GENEC & DDI & SEMEVAL \\
    \midrule
    \texttt{batch size} & 4 & 4 & 4 & 4 \\
    \texttt{max length} & 256 & 256 & 256 & 256 \\
    \texttt{optimizer} & \texttt{AdamW} & \texttt{AdamW} & \texttt{AdamW} & \texttt{AdamW} \\
    \texttt{lr} & 3e-5 & 3e-5 & 3e-5 & 3e-5 \\
    \texttt{gradient acc. steps} & 4 & 4 & 4 & 4 \\
    \texttt{warmup proportion} & 0.06 & 0.06 & 0.06 & 0.06 \\
    \texttt{max grad norm} & 1  & 1  & 1  & 1  \\
    \texttt{epoch} & 30 & 10 & 15 & 20 \\
     \bottomrule

  \end{tabular}
\end{table}

\section{Seq2SeqLM hyperparameters settings}
In all evaluation experiments, the \texttt{t5-base}\footnote{\url{https://huggingface.co/google-t5/t5-base}} model from the \texttt{huggingface} models library were used for training all datasets. All models are implemented in Python with \texttt{Pytorch}\footnote{\url{https://pytorch.org/}} and \texttt{Transfomer}\footnote{\url{https://huggingface.co/docs/transformers/en/index}} library. The random seed of 203 is set for all experiments. 
Table~\ref{tab:t5} summarizes the hyperparameter values for fine-tuning the \texttt{t5-base} model experiments. 

\begin{table}[h!]
\centering
  \caption{\texttt{t5-base} hyperparameter values.}
  \label{tab:t5}
  \begin{tabular}{lcccc}
    \toprule
     \textit{hyperparameter}     & COMAGC & GENEC & DDI & SEMEVAL \\
    \midrule
    \texttt{batch size} & 8 & 4 & 8 & 8 \\
    \texttt{max length} & 256 & 256 & 256 & 256 \\
    \texttt{optimizer} & \texttt{AdamW} & \texttt{AdamW} & \texttt{AdamW} & \texttt{AdamW} \\
    \texttt{lr} & 3e-3 & 3e-4 & 3e-4 & 3e-4 \\
    \texttt{gradient acc. steps} & 4 & 4 & 4 & 4 \\
    \texttt{warmup proportion} & 0.06 & 0.06 & 0.06 & 0.06 \\
    \texttt{max grad norm} & 1  & 1  & 1  & 1  \\
    \texttt{epoch} & 30 & 20 & 20 & 20 \\
    \bottomrule
  \end{tabular}
\end{table}

\section{Querying the KGs}
\label{ap1}
We access the \textbf{Hetionet} KG through its official public Neo4j API\footnote{\url{bolt://neo4j.het.io}}, with \texttt{neo4j.v1} Python library. Neo4j\footnote{\url{https://neo4j.com/}} is a third-party graph database that supports the Cypher language for querying and visualizing a knowledge graph. Hetionet also provides a public Neo4j browser app\footnote{\url{https://neo4j.het.io/browser/}}.

As for the \textbf{Wikidata}, we access the KG through its official public SPARQL endpoint\footnote{\url{https://query.wikidata.org/sparql}}, with \texttt{SPARQLWrapper} Python library. We employ the official wikidata API (e.g., \texttt{wbsearchentities} and \texttt{wbgetentities} functions) for extracting the Wikidata IDs for all variable pairs.

On Hetionet, we query up to 4 hops for extracting the KGs structures. However, for Wikidata, we query up to one hops to extract the KG structures, constrained by its huge sizes. To train a robust models that generalize well given any KG structures, we opted to not optimize the content from the KG structures to be included in the prompt. For instance, when there are more than $m$ metapaths for a pair, we randomly select $m$ of them, $m$ being a \textit{hyperparameter} of the number of metapaths to be included as prompt.

In all experiments, we include up to the following: 4 neighbors nodes, 5 common neighbors nodes, and 1 metapath, to be included in the prompt. These values are selected based on hyperparameters setting giving the best results in our experiments.

\section{Prompt hyperparameters and example}
(to-be-added)

\bibliographystyle{splncs04}
\bibliography{ref2}